\pdfoutput=1
\documentclass[final,5p,times,twocolumn]{elsarticle} 
\biboptions{numbers,sort&compress} 

\usepackage{amssymb}
\usepackage{lipsum}
\usepackage{amsmath}
\usepackage{float}
\usepackage{tabularx}   
\usepackage{booktabs}   
\usepackage{caption}    

\usepackage{graphicx}
\usepackage{float}            
\usepackage[section]{placeins}
\usepackage{hyperref}
\usepackage{subcaption}       
\usepackage{booktabs}
\usepackage{siunitx}
\usepackage{caption}
\journal{Astronomy $\&$ Computing}

\begin{document}

\begin{frontmatter}



\title{\textsc{\textsc{V-SenseDrive}}: A Privacy-Preserving Road Video and In-Vehicle Sensor Fusion Framework for Road Safety \& Driver Behaviour Modelling}

\author[a]{Muhammad Naveed}
\author[a]{Nazia Perwaiz}
\author[a]{Sidra Sultana}
\author[a]{Mohaira Ahmad}
\author[a]{Muhammad Moazam Fraz}

\affiliation[a]{organization={School of Electrical Engineering and Computer Science (SEECS), National University of Sciences and Technology (NUST)},
            city={Islamabad},
            country={Pakistan}}

\begin{abstract}
Road traffic accidents remain a major public health and economic challenge, particularly in countries with heterogeneous road conditions, mixed traffic flow, and variable driving discipline, such as Pakistan. Reliable detection of unsafe driving behaviours is a prerequisite for improving road safety, enabling advanced driver assistance systems (ADAS), and supporting data driven decisions in insurance and fleet management. Although several driving datasets have been developed internationally, most originate from controlled environments in developed countries, with limited representation of the cultural, infrastructural, and behavioural diversity observed in emerging economies. This Article presents \textsc{V-SenseDrive}, the first multimodal driver behaviour dataset collected entirely within the Pakistani driving environment. The dataset captures local traffic dynamics, varied road surface conditions, and diverse vehicle types, offering a realistic foundation for behaviour recognition research. \textsc{V-SenseDrive} combines smartphone based inertial and GPS sensor data with synchronized road facing video to record three target driving behaviors (normal, aggressive, and risky) on multiple types of roads, including urban arterials, secondary roads, and motorways. Data was gathered using a custom Android application designed to capture high frequency accelerometer, gyroscope, and GPS streams alongside continuous video, with all sources precisely time aligned to enable multimodal analysis. The focus of this work is on the data acquisition process, covering participant selection, driving scenarios, route planning, environmental considerations, and sensor video synchronization techniques. The dataset is structured into raw, processed, and semantic layers, ensuring adaptability for future research in driver behaviour classification, traffic safety analysis, and ADAS development. By representing real world driving in Pakistan, \textsc{V-SenseDrive} fills a critical gap in the global landscape of driver behaviour datasets and lays the groundwork for context aware intelligent transportation solutions.
\end{abstract}



\begin{keyword}
Advanced Driver Assistance Systems (ADAS), Explainable AI, Inertial Measurement Unit (IMU), Low and Middle Income Countries (LMICs), \textsc{V-SenseDrive}.



\end{keyword}

\end{frontmatter}




\section{Introduction}
Road-traffic injuries remain one of the most urgent public health and socioeconomic issues worldwide. According to the \cite{WHO2023RoadTrafficInjuries}, road accidents claim an estimated 1.19 million lives annually and injure between 20 and 50 million people, often resulting in permanent disability and long term economic hardship. Low and middle income countries (LMICs) are disproportionately affected, accounting for more than 90\% of global deaths despite owning just over half of the world’s motor vehicles. The economic cost of road traffic accidents is also substantial, consuming between 1\% and 3\% of GDP in most LMICs as per \cite{WorldBank2018RoadSafety}. Pakistan presents a particularly acute case within this global crisis. Rapid urbanization, unregulated traffic growth, heterogeneous vehicle fleets, and varied road infrastructure from congested city streets to underdeveloped rural roads create a highly complex driving environment. This environment is further complicated by mixed traffic flows that involve cars, motorcycles, rickshaws, buses, trucks, and pedestrians sharing the same space, often without strict lane discipline as per \cite{Shah2021UrbanMobility}. In such conditions, unsafe driving behaviours, ranging from excessive speeding and abrupt lane changes to aggressive overtaking and risky close following, can quickly lead to collisions and casualties.

Therefore, accurate recognition and monitoring of driver behaviour is essential to improve road safety, inform policy, enable insurance risk assessment, and develop advanced driver assistance systems (ADAS) tailored to the realities of Pakistan's roads as per \cite{PJHSS2024}. However, such recognition systems are critically dependent on high quality region specific datasets that capture both the sensory and contextual aspects of driving in real world conditions. Although previous efforts, such as the \cite{romera2016need}, have demonstrated the feasibility of smartphone based driver behaviour monitoring, these datasets were collected in controlled, high income environments with more uniform traffic laws, better infrastructure, and predictable driving patterns. Their geographic, behavioural, and cultural limitations make them suboptimal for training models to operate effectively in the complex and often chaotic driving scenarios of Pakistan.

To bridge this gap, we focus on the systematic acquisition of a Pakistan-specific, multimodal driver behaviour dataset, V-SenseDrive. The dataset was generated through a purpose-designed acquisition pipeline that integrates smartphone-based inertial and positional sensors such as the accelerometer, gyroscope, and GPS with road-facing video streams to capture the surrounding environment. A key feature of this pipeline is the precise temporal synchronization between sensor readings and visual data, which ensures that environmental context and motion dynamics can be jointly analysed. This multimodal alignment enables advanced fusion techniques, thereby providing a robust foundation for developing and evaluating driver behaviour models tailored to the Pakistani context.

The acquisition process was carefully designed to:
\begin{enumerate}
    \item \textbf{Reflect real-world conditions}: Data was collected across diverse road types (motorways, urban streets, rural roads) and traffic densities, under varying lighting and weather conditions.
    \item \textbf{Capture authentic behaviours}: Driving sessions included naturalistic instances of normal, aggressive, and risky driving, observed during everyday journeys rather than being fully simulated.
    \item \textbf{Support sensor vision alignment}: All modalities were synchronised to facilitate subsequent fusion between motion signals and semantic scene understanding.
\end{enumerate}

In addition to raw sensor and video data, the acquisition framework incorporates metadata such as road category, traffic density, and environmental conditions, enabling richer downstream analysis. This emphasis on methodical data acquisition ensures that \textsc{V-SenseDrive} not only serves as a high value dataset for model training but also as a reference framework for future region specific driver monitoring studies.

The contributions of this work, from a data acquisition perspective, are as follows:
\begin{itemize}
    \item A regionally tailored acquisition protocol designed for heterogeneous traffic environments.
    \item A multimodal dataset combining motion sensors, GPS, and visual context.
    \item An open methodological framework that can be adapted for similar LMIC traffic conditions.
\end{itemize}

By focusing on the \textit{how} of data acquisition, this paper not only delivers a novel dataset but also provides a replicable blueprint for capturing complex, real world driving behaviour in underrepresented traffic contexts.

\section{Related Work}
A substantial body of research has focused on creating and analysing driving behaviour datasets to support applications in driver safety assessment, behaviour recognition, and intelligent transportation systems. Existing datasets vary in their sensing modalities, environmental diversity, behaviour taxonomies, and geographic contexts. This section reviews representative contributions, grouped by sensing approaches and application focus, to position the novelty of the \textsc{V-SenseDrive} dataset.

\subsection{Smartphone-Based Datasets}
The proliferation of mobile devices has enabled low cost, easily deployable driving behaviour sensing. The UAH DriveSet dataset~\cite{romera2016need} demonstrated the feasibility of smartphone based detection of normal, drowsy, and aggressive behaviours, combining inertial, GPS, and video data from six drivers in Spain. While influential, it was geographically restricted and covered only motorway and secondary roads. Similarly, Wawage and Deshpande~\cite{wawage2022smartphone} presented a smartphone sensor dataset capturing accelerometer, gyroscope, magnetometer, and GPS readings for driver behaviour analysis, although without integrated video or diverse environmental contexts. Cojocaru and Popescu~\cite{cojocaru2022building} also leveraged Android devices to build a labelled driving dataset focused on calm versus aggressive styles, collected under controlled urban settings. These efforts highlight the potential of commodity devices for behavioural monitoring but show limited representation of varied traffic, weather, and cultural driving norms.

\subsection{Multimodal Sensor and Video Datasets}
High-fidelity behavioural analysis often benefits from combining vehicular telemetry with video and additional sensor streams. Tao et al.~\cite{tao2024multimodal} introduced a multimodal physiological dataset linking driver biometrics with vehicle signals and contextual video, enabling fine-grained state monitoring. Huang et al.~\cite{huang2021driving} used large-scale naturalistic driving data with inverse reinforcement learning to model decision making processes, emphasising lane-change and car-following behaviours. In survey work, Adhikari~\cite{adhikari2023using} comprehensively reviewed visual and vehicular sensing modalities for driver behaviour analysis, identifying the challenges of sensor fusion, synchronisation, and scalability issues the \textsc{V-SenseDrive} design directly addresses.

\subsection{Driving behaviour Modelling and Machine Learning Applications}
Recent works have also examined advanced modelling techniques. Talebloo et al.~\cite{talebloo2021deep} proposed deep learning architectures for aggressive driving detection, integrating temporal and spatial sensor features. Peppes et al.~\cite{peppes2021driving} evaluated machine and deep learning methods on continuous vehicular data streams, showing the role of real time analytics for safety critical applications. Cojocaru et al.~\cite{cojocaru2022driver} further explored deep learning based behavioural recognition using multimodal inputs, but their experiments were constrained to limited environments.

\subsection{Simulation, Crash Risk Studies, and IoT Datasets}
In contrast to naturalistic data collection, simulated or reconstructed driving datasets can target rare or dangerous scenarios. Scanlon et al.~\cite{scanlon2021waymo} analysed autonomous vehicle behaviour in reconstructed fatal crash scenarios, providing insights into safety critical decision making. Ma et al.~\cite{ma2023prediction} modelled freeway crash likelihood by incorporating risky driving behaviours as predictive variables, demonstrating the value of behavioural features in accident risk assessment.

\renewcommand{\arraystretch}{1.8} 
\begin{table*}[htbp]
\centering
\caption{Comparison of existing open driving behavior datasets including the proposed \textsc{\textsc{V-SenseDrive}}}
\resizebox{\textwidth}{!}{%
\begin{tabular}{p{2.8cm}p{6cm}p{1.5cm}p{2cm}p{3.5cm}p{2cm}p{3.5cm}}
\hline
\textbf{Dataset / Study} & \textbf{Devices \& Sensors Used} & \textbf{Volume} & \textbf{Duration} & \textbf{Parameter Variations} & \textbf{Drivers} & \textbf{Use Cases} \\
\hline
\multicolumn{7}{l}{\textbf{(A) Sensor-only datasets}} \\
\hline
Smartphone Sensor Dataset \cite{wawage2022smartphone} & Smartphone: accelerometer, gyroscope, proximity & CSV only & 5--20 km trips & Normal vs risky events & -- & Aggression \\
Aggressive Driving \cite{popescu2022kaggle} (Kaggle) & Smartphone accelerometer + gyroscope & -- & -- & Slow, normal, aggressive & -- & Aggression \\
Revitsone-5Class \cite{revitsone2022} & Smartphone accelerometer + gyroscope & -- & -- & 5 classes: safe, texting, calling, turning, other & -- & Inattention, Behavior \\
Driving Behaviour \cite{yuksel2020mendeley} & Raspberry Pi + MPU6050 (acc/gyro) & -- & 2 weeks & Sudden accel, turns, braking & 3 male drivers (27--37y) & Aggression \\
DriverBehaviorDataset \cite{driverbehaviordatasetgithub} (GitHub) & Smartphone accelerometer + gyroscope + magnetometer & -- & $\sim$4 trips $\times$ 13 min & Braking, accel, turns, lane change & 2 (15y exp) & Aggression \\
Aggressive Driving IoT \cite{singh2022vehicle} & IoT: GPS, accelerometer, gyroscope, OBD & 26 MB (CSV) & -- & Aggressive vs non-aggressive & Fleet-based & Aggression (IoT telematics) \\
\hline
\multicolumn{7}{l}{\textbf{(B) Face-only / driver-focused datasets}} \\
\hline
NTHU Drowsy Driver \cite{weng2016driver} & IR camera (simulated) & -- & 9h 30m & Illumination, glasses/no glasses & 18 (8F, 10M) & Inattention \\
AUC-DDD \cite{eraqi2019driver} & Driver side profile images & -- & N/A & 10 task-specific scenarios & V1: 31; V2: 44 & Inattention \\
RoCHI \cite{cojocaru2022building1} (2022) & Eye-gaze tracker + head-pose camera & -- & 10h & Distracted (phone, fatigue) & 15 & Inattention \\
\hline
\multicolumn{7}{l}{\textbf{(C) Road video only datasets}} \\
\hline
YawDD \cite{e1qm-hb90-20} & RGB camera (rear-mirror) + dashcam & 4.9 GB & 2--3h & Weather: sunny, cloudy, rainy & 107 (50F, 57M) & Inattention \\
\hline
\multicolumn{7}{l}{\textbf{(D) Road + sensors (smartphone/dashcam-based)}} \\
\hline
UAH-DriveSet (2016) \cite{romera2016need} & Smartphone sensors (GPS, accelerometer, gyroscope) + front camera & 3.3 GB & $\sim$4h 20m & Motorway \& secondary road; normal, drowsy, aggressive & 6 (1F, 5M) & Aggression \\
\textsc{\textsc{V-SenseDrive}} (Proposed) & Multimodal: Dashcam (road scenes) + IMU + GPS & 85.5 GB & 16h (field) & Mixed traffic, aggressive, distracted, intoxicated & 4 & Aggression, Inattention, Intoxication \\
Building Driving Behaviour \cite{cojocaru2022building} (2022) & Smartphones + dashcams + CAN bus & 500 GB & 100h & City vs highway; normal vs risky & 25 & Aggression \\
Sensors (2021) & Dashcam video + IMU (acc, gyro) + GPS & -- & $\sim$50h & Urban, highway, mixed traffic & 20+ & Aggression, Inattention \\
\hline
\multicolumn{7}{l}{\textbf{(E) Complex multimodal datasets (road + face + vehicle signals)}} \\
\hline
SHRP2 NDS \cite{ahmed2018driver} & 4 cameras (scene, face, lap, hand), GPS, radar, CAN-bus, cellphone & 2 PB & 2 years & Varied surface, weather, lighting & $>$3000 & Inattention, Aggression \\
Brain4Cars \cite{jain2016brain4cars} & Cameras (front \& rear), CAN bus & 104h & -- & Lane changes, turns, merging maneuvers & 100+ & Inattention (maneuver prediction) \\
100-Cars Study \cite{phase2006100} & Driver + road-facing cameras, GPS, radar, cellphone & 6.4 TB & 43{,}000h & Varied surface, weather, lighting & 109 (43F, 66M) & Aggression, Inattention \\
UTDrive \cite{liu2021review} (2021) & Video (road + cabin) + CAN + GPS + IMU + face/eye monitoring & $\sim$50 GB & $\sim$12h & Urban \& highway; distracted vs attentive & 60+ & Inattention, Aggression \\
Drive\&Act \cite{martin2019drive} (2019) & Multi-view RGB + depth + thermal + skeleton + IMUs (12 modalities) & 3 TB & 147 drivers × sessions & Secondary tasks: phone, eating, fatigue & 147 & Inattention, Secondary Tasks, Behavior \\
DBNet \cite{chen2019dbnet} (2019, CVPR) & Velodyne LiDAR + dashcam video + vehicle telemetry (speed, steering) & -- & -- & Speed, steering, depth trajectories & -- & Behavior learning for AVs \\
Inverse RL \cite{huang2021driving} & Naturalistic vehicle sensors + physiological (EEG/ECG/HR) & -- & -- & Aggressive vs normal trajectories & 30+ & Aggression, Intoxication \\
PJHSS \cite{irfan2024road} (2024) & Surveys + simulator + vehicle sensors & -- & 8h (simulated runs) & Intoxication vs normal driving & 40 (local drivers) & Intoxication \\
hciLab Driving Dataset \cite{hcilab2013} & Physiological sensors + video rating & $\sim$38 MB & Real-world experiment & Road type workload (exits, ramps) & 10 & Workload / Human Factors \\
\hline
\end{tabular}%
}
\label{tab:datasets}
\end{table*}

Beyond traditional simulation or smartphone only datasets, the \textit{Aggressive Driving behaviour IoT Data}~\cite{shuvo2024aggressive} represents a recent effort to capture real time driving behaviour via an Internet of Things (IoT) framework. It integrates GPS, accelerometer, gyroscope, and onboard diagnostics data, alongside binary event indicators (e.g., harsh braking, harsh acceleration, sharp turns) and a categorical aggressive behaviour label. This dataset enables predictive modelling of aggressive driving patterns and supports applications in traffic safety, insurance risk profiling, and intelligent transportation systems.

\subsection{Gaps in Literature and Contribution of \textsc{V-SenseDrive}}

Despite notable advancements in driver behavior analysis, existing datasets exhibit several critical limitations. First, there is a lack of representation from low and middle income countries (LMICs), where traffic conditions are markedly different due to infrastructural, cultural, and enforcement disparities. Second, few datasets offer concurrent high resolution video and synchronized smartphone sensor data with frame level alignment, limiting their applicability in fine grained multimodal modeling. Third, existing behavioral taxonomies often lack cross cultural applicability, and many datasets do not capture scripted, yet contextually realistic, driving behaviors across diverse road types and conditions.

The \textsc{V-SenseDrive} dataset addresses these challenges through a novel multimodal acquisition framework. It integrates high frequency inertial measurements (accelerometer, gyroscope, magnetometer) and GPS data with synchronized 1080p road facing video, all captured using commodity Android smartphones. Data were collected across urban streets, secondary roads, and expressways in Pakistan a region largely absent from publicly available datasets. Driving behaviors were systematically enacted and categorized into three classes Normal, Aggressive, and Risky based on an adapted and culturally validated taxonomy derived from UAH DriveSet~\cite{romera2016need}. Manual verification ensured label accuracy, while the dataset’s diversity in road context and driving conditions enhances its generalizability and modeling potential.

\subsection{Significance of the Proposed Dataset}

The \textit{\textsc{V-SenseDrive}} dataset constitutes a significant advancement in the domain of driver behavior research, particularly within the context of LMICs. Although several high fidelity data sets such as SHRP2 NDS~\cite{ahmed2018driver}, Brain4Cars~\cite{jain2016brain4cars}, and Drive\&Act~\cite{martin2019drive} have substantially contributed to understanding driver behavior, they predominantly originate from high income countries and rely on expensive proprietary sensing systems, including LiDAR, radar, CAN bus signals, and biometric data. Such dependence not only increases the cost and complexity of data acquisition but also limits the feasibility of replicating these systems in resource constrained regions. \textsc{V-SenseDrive} overcomes these barriers by leveraging widely available Android smartphones as the sole data acquisition platform. The sensing suite includes 1080p/30fps video along with GPS, accelerometer, gyroscope, and magnetometer data captured without the use of any specialized or intrusive equipment. This approach democratizes data collection and enables affordable behavioral monitoring in developing countries, supporting applications in policy making, telematics, insurance, and ADAS development.

Moreover, \textsc{V-SenseDrive} introduces a robust synchronization mechanism that aligns video frames with sensor readings at a fine temporal resolution. This integration allows for precise multimodal fusion, which is often lacking in publicly available smartphone based datasets~\cite{wawage2022smartphone, popescu2022kaggle}. The dataset thereby supports advanced deep learning architectures such as BiLSTM and sensor vision fusion networks capable of leveraging both environmental context and dynamic motion cues. In addition to its technical strengths, \textsc{V-SenseDrive} provides critical geographical and cultural relevance. The data set captures the unstructured, heterogeneous, and often unpredictable nature of South Asian traffic characterized by mixed traffic users, informal traffic rules, and variable driving etiquette. Such behavioral variance is essential for developing globally generalizable and culturally adaptive driver monitoring systems. By offering a low cost, scalable, and interpretable dataset grounded in real world LMIC traffic conditions, \textsc{V-SenseDrive} fills a key gap in current literature and sets a new precedent for inclusive, multimodal driver behavior research.

\section{Setup and Methodology}

This study employs a comprehensive, modular methodology to develop \textbf{\textsc{V-SenseDrive}}, an intelligent, multimodal driver behaviour classification and video annotation system designed for operation in real world Pakistani traffic conditions. The methodological framework spans the entire data lifecycle, beginning with the acquisition of multimodal data from in-vehicle sensors and forward-facing video, followed by precise temporal synchronisation, signal preprocessing, feature engineering, and rule based behaviour labelling. This ensures that the resulting dataset is both analytically robust and suitable for training advanced deep learning models for behaviour recognition and scene interpretation.

The core objective of \textsc{V-SenseDrive} is to classify driver behaviour into three distinct and operationally meaningful categories:
\begin{enumerate}
    \item \textbf{Normal driving:} Calm, lawful driving patterns that maintain safe speeds and following distances.
    \item \textbf{Aggressive driving:} Assertive and speed focused manoeuvres, including rapid acceleration, sharp lane changes, and speeding.
    \item \textbf{Risky driving:} Potentially hazardous behaviour characterised by unstable turns, late braking, lane drifting, and erratic speed variation.
\end{enumerate}

\subsection{Data Acquisition}

The first stage of this work involved the systematic collection of synchronized multimodal data to build a robust foundation for driver behaviour analysis. Since the accuracy of downstream modeling and the interpretability of results depend directly on the quality of raw inputs, a carefully designed acquisition strategy was implemented. In this study, naturalistic driving trips were recorded under diverse real-world conditions using two off-the-shelf smartphones. One device was dedicated to logging high frequency inertial and positional signals, while the other continuously recorded forward-facing road video. This dual modality approach ensured that both the internal dynamics of vehicle motion and the external driving context were captured, allowing richer behavioural insights and enabling the development of an explainable driver behaviour classification system.

\subsubsection{Dual Modality Sensing Framework}

Reliable multimodal data acquisition formed the foundation of the \textsc{V-SenseDrive} dataset. To balance affordability, replicability, and high fidelity, a dual smartphone sensing framework was adopted. This configuration allowed the simultaneous recording of two complementary modalities: (i) inertial and geolocation signals capturing the kinematics of vehicle motion, and (ii) forward facing video capturing the external road environment. By integrating these two perspectives, the system provided both quantitative measurements of driving dynamics and qualitative contextual cues from the surrounding traffic scene.

\begin{figure}[H]
    \centering
    \includegraphics[width=\linewidth]{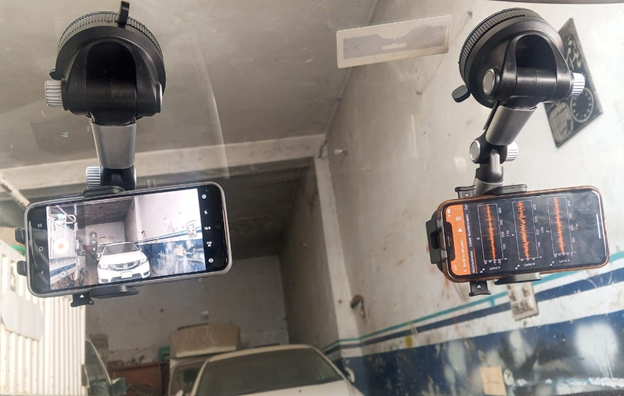} 
    \caption{Dual Smartphone Data Acquisition Setup}
    \label{fig:Data Collection Setup}
\end{figure}

Figure~\ref{fig:Data Collection Setup} illustrates the dual smartphone acquisition setup used in this study. The video smartphone was mounted centrally on the dashboard to maximize the field of view of the road environment, while the sensor smartphone was secured on the windshield using a fixed holder to minimize vibration artifacts and maintain clear line of sight for GPS signals. In this configuration, the left device continuously captured forward facing video, and the right device simultaneously streamed multi sensor signals via the Phyphox application. Mounting both devices in stable positions ensured synchronized and reliable recordings under diverse real world driving conditions, including urban congestion, highway travel, and mixed traffic scenarios, thereby providing a comprehensive foundation for driver behaviour analysis.

\subsubsection{Sensor Device and Recorded Modalities}

The primary sensing platform was an \textbf{Apple iPhone X} (Model iPhone10,3), chosen for its high-quality Inertial Measurement Unit (IMU) sensors and reliable GPS hardware. Data acquisition was performed using the \textit{Phyphox} mobile application (version 1.1.16) developed by RWTH Aachen University, which enables high-frequency logging with UNIX timestamp resolution. Sensor sampling was configured at approximately 25\,Hz, providing sufficient temporal granularity to capture short-duration driving events such as harsh braking, swerving, or rapid acceleration. The recorded modalities encompassed a comprehensive suite of signals that together reflected both vehicle dynamics and environmental context. The accelerometer measured total linear acceleration across three axes, capturing speed fluctuations and braking intensity, while the linear accelerometer, obtained by filtering gravitational components, isolated pure translational motion for detecting manoeuvres such as lane changes and cornering. The gyroscope recorded angular velocity along three axes, facilitating recognition of steering behaviour, sharp turns, and vehicle drift. The magnetometer logged ambient magnetic field strength to provide compass-based directional cues and supported orientation estimation. Through sensor fusion of accelerometer, gyroscope, and magnetometer readings, orientation angles (pitch and yaw) were derived, enabling analysis of longitudinal tilt (e.g., downhill braking) and lateral heading changes. Finally, the GPS module contributed latitude, longitude, altitude, velocity, and bearing, accompanied by positional accuracy estimates, thereby situating vehicle dynamics within road topology and traffic context. Collectively, these synchronized multimodal streams provided a rich foundation for the naturalistic analysis of driver behaviour.

\subsubsection{Video Capture}

In parallel, a \textbf{Samsung Galaxy A16} smartphone was dedicated to recording the forward-facing road environment. The device was mounted on the dashboard with a stable holder and configured to record video in high definition (1920$\times$1080 pixels) at 30\,frames per second (FPS). This ensured that critical environmental cues including lane markings, surrounding vehicles, traffic signs, and pedestrian movements were captured with sufficient clarity. The frame rate of 30\,FPS was chosen as a balance between temporal smoothness (to capture transient events such as overtakes) and storage efficiency, resulting in a total video data set that exceeds 85\,GB in all recording sessions.

The video and sensor data streams were captured simultaneously during naturalistic driving trips across diverse conditions, including urban arterial roads, congested intersections, highways, and residential streets. Environmental diversity was deliberately included to improve the generalizability of the dataset and expose the models to heterogeneous driving contexts.

\subsubsection{Importance of Dual-Modality Acquisition}

The simultaneous acquisition of sensor and video data was fundamental for constructing an interpretable driver behaviour analysis framework, as reliance on a single modality would inevitably introduce blind spots. Inertial and positional sensors (accelerometer, gyroscope, magnetometer, GPS) capture internal vehicle dynamics such as acceleration bursts, yaw shifts, or sudden braking, but lack contextual awareness of external factors. This can lead to misinterpretation, for instance classifying defensive braking as reckless behaviour. Conversely, forward-facing video provides rich contextual cues about road layout, traffic participants, signage, and weather conditions, but cannot quantify internal vehicle inputs such as braking intensity or lateral forces. By synchronizing both modalities at the frame level, \textsc{V-SenseDrive} integrates precision with context: sensors reveal \textit{what} manoeuvres occurred, while video clarifies \textit{why} they happened. This synergy reduces ambiguity, enhances interpretability, and improves trust in downstream model outputs, forming a robust foundation for multimodal fusion and explainable AI in driver behaviour recognition.  

\begin{table}[H]
  \centering
  \setlength{\tabcolsep}{6pt}
  \renewcommand{\arraystretch}{1.15}
  \small
  \caption{Specifications and synchronization methods for dual-modality data acquisition.}
  \label{tab:sensor_video_setup}
  \begin{tabularx}{\columnwidth}{@{} l X @{}}
    \toprule
    \textbf{Component} & \textbf{Details} \\
    \midrule
    Sensor Device        & iPhone X with Phyphox app (mounted horizontally on windscreen) \\
    Sensor Types         & Accelerometer, Linear Acceleration, Gyroscope, Magnetometer, Orientation, GPS \\
    Sampling Rates       & IMU: $\sim$50 Hz; GPS: 1 Hz \\
    Video Device         & Samsung Galaxy A16, forward-facing, 1080p at 30 FPS \\
    Mounting             & Adjacent to iPhone X, horizontally aligned for consistent perspective \\
    Synchronization      & Manual cues + UNIX timestamp normalization \\
    Alignment Method     & Linear interpolation + \texttt{merge\_asof()} for frame-level fusion \\
    Output               & Frame-aligned sensor dataset with synchronized video stream \\
    \bottomrule
  \end{tabularx}
\end{table}

This configuration ensured precise temporal alignment between environmental context and measured vehicle dynamics, minimizing artefacts through stable mounting and time synchronization before each session. The resulting dataset offers high fidelity, multimodal recordings that enable robust and interpretable analysis of naturalistic driving behaviours.

\begin{figure}[H]
    \centering
    \includegraphics[width=\linewidth]{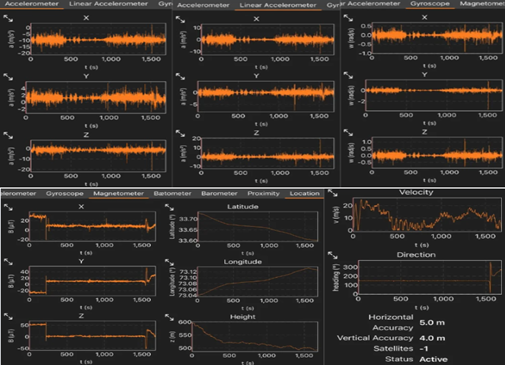} 
    \caption{Real-Time Multi-Sensor Data Streams During Driving}
    \label{fig:realtime_sensor_streams}
\end{figure}

Figure~\ref{fig:realtime_sensor_streams} presents representative multi-axis sensor outputs, including accelerometer, gyroscope, magnetometer, barometer, GPS, and derived motion parameters. The plots show continuous sampling of kinematic and positional variables, enabling fine grained analysis of vehicle dynamics and driver behaviour patterns.

\subsubsection{Road Type and Route Characteristics}
Data was collected across two road categories to maximize environmental diversity:

\vspace{1.0\baselineskip} 
\noindent
\begin{minipage}{\columnwidth}
  \centering
  \setlength{\tabcolsep}{6pt}
  \renewcommand{\arraystretch}{1.2}
  \small
  \captionof{table}{Driving Routes, Distances, and Road Environment Characteristics}
  \label{tab:road_network_coverage}
  \begin{tabular}{@{} p{1.2cm} p{2.3cm} p{1.0cm} p{3.0cm} @{}} 
    \toprule
    \textbf{Road Type} & \textbf{Route} & \textbf{Distance (km)} & \textbf{Characteristics} \\
    \midrule
    \textbf{Secondary} & Sirsyed Chowk $\leftrightarrow$ Koral Chowk & $\sim$13.0 & Mixed traffic, irregular lane markings, high pedestrian density, heterogeneous vehicles. \\
    \textbf{Highway} & Koral Chowk $\leftrightarrow$ Faisal Mosque & $\sim$34.2 & Higher speeds, variable lane discipline, heavy and light vehicles, occasional stationary obstacles. \\
    \bottomrule
  \end{tabular}
\end{minipage}
\vspace{1.0\baselineskip} 

Table~\ref{tab:road_network_coverage} outlines the driving routes, distances, and contextual conditions considered. These routes were selected to encompass diverse road environments, ranging from congested urban secondary roads to higher speed highways, enabling the capture of varied driving behaviours. Each route was traversed in both directions (forward and return), at multiple times of day (morning, afternoon, evening, night) and under varied weather conditions (clear, strong sunlight, heavy rain), ensuring a robust dataset.

\subsubsection{behavioural Protocols}
Three driving behaviour modes, adapted from the UAH-DriveSet taxonomy and contextualized for Pakistani roads, were enacted:

\begin{itemize}
    \item \textbf{Normal Driving:} Smooth acceleration, gentle braking, lane discipline, safe following distances, controlled turns, and consistent speed.
    \item \textbf{Aggressive Driving:} Excessive speeding (+50 km/h urban, +70 km/h highway), rapid acceleration, hard braking, frequent lane changes, tailgating, and unsafe mergers.
    \item \textbf{Risky Driving:} Lane drifting, inconsistent speeds, late braking, unstable turns, hydroplaning, driving in reduced visibility, emergency evasive maneuvers, and near-miss incidents.
\end{itemize}

All non normal behaviours were scripted and closely monitored to ensure safety. Each behavioural mode was repeated across varying traffic densities and road conditions to capture a comprehensive range of scenarios. This systematic approach ensured balanced data representation for all categories, enabling robust model training and evaluation.

\subsubsection{Participant and Vehicle Diversity}
Data was collected from four drivers operating vehicles with differing transmissions, engine capacities, and mechanical characteristics:

\vspace{1.0\baselineskip} 
\noindent
\begin{minipage}{\columnwidth}
  \centering
  \setlength{\tabcolsep}{4pt}     
  \renewcommand{\arraystretch}{1.2}
  \scriptsize                     
  \captionof{table}{Vehicle, Participant, and Driving Condition Diversity}
  \label{tab:participant_vehicle}
  \begin{tabular}{@{} p{1.7cm} p{1.2cm} p{1.0cm} p{2.0cm} p{2.1cm} @{}} 
    \toprule
    \textbf{Vehicle} & \textbf{Transmission} & \textbf{Engine} & \textbf{behaviour Types} & \textbf{Time/Weather} \\
    \midrule
    \textbf{Suzuki Mehran VXR}\\(2003) 
      & Manual 
      & 796 cc 
      & Normal, Aggressive, Risky 
      & Morning clear; afternoon hot; night rain \\
    \textbf{Suzuki Wagon R}\\(2017) 
      & Manual 
      & 998 cc 
      & Normal, Aggressive, Risky 
      & Early morning clear; later sunlight \\
    \textbf{Toyota Corolla GLi}\\(2012) 
      & Manual 
      & 1299 cc 
      & Normal, Aggressive, Risky (drowsy) 
      & Evening \& night clear \\
    \textbf{Honda City}\\(2019) 
      & Automatic 
      & 1500 cc 
      & Normal, Aggressive, Risky 
      & Evening \& night clear \\
    \bottomrule
  \end{tabular}
\end{minipage}
\vspace{1.0\baselineskip} 

Table~\ref{tab:participant_vehicle} highlights the variety of vehicles, transmission types, and operating conditions included in the dataset. This diversity supports broader generalization of driving behaviour models across different driver profiles, vehicle classes, and environmental contexts.

\subsubsection{Synchronization Protocol}
Because the Phyphox application and the video recording operated on separate devices, a manual synchronization procedure was implemented to ensure precise temporal alignment between the two data streams. Both devices were initiated simultaneously before departure, after which a distinct visual and audible cue, such as a hand clap, was performed at the start and end of each trip to serve as a common reference point. During post processing, UTC timestamps from the Phyphox logs were matched with the corresponding video frame indices, and linear interpolation was applied to map each sensor reading to its corresponding video frame. This process enabled frame level fusion of physical motion data and environmental visual context, ensuring multimodal integration for accurate behaviour analysis.

\subsection{Data Preprocessing and Feature Engineering}

\subsubsection{Timestamp Harmonization}
The raw sensor data, exported from Phyphox as an Excel workbook containing separate sheets for each modality, underwent a timestamp harmonization process to enable seamless integration. All \texttt{Time (s)} columns were first renamed to prevent naming conflicts, and a single global timestamp was created by offsetting from the known trip start time. Using this unified temporal reference, the individual sensor streams were merged using the Pandas \texttt{merge\_asof} function, applying a maximum tolerance of one second to maintain precise chronological alignment across all modalities.

\subsubsection{GPS Speed Computation}
GPS derived speeds were calculated by first determining the distance between consecutive coordinate pairs using the \texttt{geopy.geodesic} method, which provides high accuracy geodesic distance measurements in meters. These distances were then converted into instantaneous speeds by dividing by the corresponding time interval (in seconds) and multiplying by $3.6$ to convert from meters per second (m/s) to kilometers per hour (km/h), as expressed by:

\begin{equation}
\label{eq:speed}
\text{Speed (km/h)} = \frac{\text{Distance (m)}}{\text{Time Interval (s)}} \times 3.6
\end{equation}

This procedure ensured precise estimation of vehicle speed at each recorded GPS point, forming a crucial input for behaviour classification.

\subsubsection{Data Interpolation and Noise Reduction}
To address missing GPS attributes, a two stage imputation process was applied. The first stage employed linear interpolation to estimate intermediate values, ensuring continuity in temporal and spatial data. The second stage used forward and backward filling to replace any residual gaps. For IMU data, a low pass filter was implemented on accelerometer and gyroscope signals to suppress high frequency vibration noise. This preserved the integrity of meaningful motion patterns while eliminating artefacts caused by road surface irregularities or mounting vibrations.

\subsubsection{Derived Features}
Additional motion features were engineered to enhance behavioural detection. Jerk, defined as the rate of change of acceleration, was computed separately for each axis ($X, Y, Z$) as:

\begin{equation}
\label{eq:jerk_x}
\text{Jerk}_X(t) = \frac{\text{Acc}_X(t) - \text{Acc}_X(t-1)}{\Delta t}
\end{equation}

The overall jerk magnitude, capturing the combined effect of motion changes across all axes, was calculated using:
\begin{equation}
\label{eq:jerk_mag}
\text{Jerk}_{\text{mag}} = \sqrt{\text{Jerk}_X^2 + \text{Jerk}_Y^2 + \text{Jerk}_Z^2}
\end{equation}

In addition, pitch and yaw angles were extracted from the orientation readings to aid in detecting steering patterns, lane changes, and turning maneuvers. These engineered features provided physically interpretable indicators of driving style and formed the basis for rule based behaviour labelling.

\subsection{Rule Based behaviour Labelling}
To translate raw sensor and derived motion features into meaningful driving behaviour categories, a two tier deterministic labelling framework was developed. This framework first detects specific sub behaviours based on physically interpretable thresholds derived from kinematic and orientation data, and then maps these sub behaviours into broader primary behaviour classes. The design ensures that each label has a clear physical basis, facilitating interpretability and reproducibility in both human and machine analysis.

\subsubsection{Sub behaviour Detection}
The rule based behaviour labelling stage forms a critical link between raw sensor data and meaningful behavioural interpretation. This approach enables the systematic categorization of driving patterns based on measurable physical indicators and relies on well defined, physically interpretable thresholds to detect specific sub-behaviours such as:
\begin{itemize}
    \item \textbf{Speeding} – exceeding the speed limit by more than the defined threshold.
    \item \textbf{Sudden Acceleration} – rapid increase in velocity exceeding acceleration threshold.
    \item \textbf{Hard Braking} – rapid deceleration beyond a negative acceleration threshold.
    \item \textbf{Lane Drifting} – sustained lateral acceleration without corresponding steering correction.
    \item \textbf{Steady Driving} – consistent velocity and low jerk magnitude.
\end{itemize}

By grounding the labelling process in quantifiable kinematic parameters, the framework ensures objectivity, reproducibility, and compatibility with subsequent machine learning and statistical analysis.

\noindent
\begin{minipage}{\columnwidth}
  \centering
  \setlength{\tabcolsep}{6pt}
  \renewcommand{\arraystretch}{1.2}
  \small
  \captionof{table}{Logical Thresholds for Automated Driving Sub-behaviour Detection}
  \label{tab:sub_behaviour_conditions}
  \begin{tabular}{@{} p{3cm} p{5cm} @{}} 
    \toprule
    \textbf{Sub-behaviour} & \textbf{Logical Condition} \\
    \midrule
    Speeding & Speed $>$ 70 km/h \\
    Sudden Acceleration & Jerk\_X $>$ 25 and Accel\_X $>$ 2.5 and Pitch $<$ -0.12 \\
    Hard Braking & Jerk\_X $>$ 25 and Accel\_X $<$ -2.5 and Pitch $>$ 0.12 \\
    Lane Drifting & Gyro\_Y $>$ 0.6 and Speed between 5--60 \\
    Sharp Turn & Gyro\_Y or Gyro\_Z $>$ 1.0 \\
    Inconsistent Speed & Jerk $>$ 12 and Accel\_X $\approx$ 0 and Speed 10--40 \\
    Steady Driving & Speed between 20--50 and Jerk\_Magnitude $<$ 7 \\
    Idle & Speed $<$ 10 and Jerk $<$ 3 \\
    \bottomrule
  \end{tabular}
\end{minipage}

\vspace{1.0\baselineskip} 

Table~\ref{tab:sub_behaviour_conditions} defines the logical thresholds for detecting key sub behaviours such as speeding, sudden acceleration, and lane drifting. These conditions were formulated based on established vehicular dynamics metrics to enable consistent and automated behaviour classification.

\subsubsection{Primary behaviour Mapping}
The second tier aggregated the detected sub-behaviours into three operationally meaningful primary categories. \textit{Aggressive driving} encompasses speeding, sudden acceleration, hard braking, and lane drifting behaviours indicative of assertive, speed focused, and forceful maneuvers. \textit{Risky driving} included patterns such as unstable turns, late braking, and inconsistent speeds, all of which have a heightened probability of leading to unsafe situations. \textit{Normal driving} captured steady driving, smooth acceleration, gentle braking, and idle periods, representing compliant and safe road behaviour. This hierarchical mapping preserved interpretability while enabling behaviour analysis at multiple levels of granularity.

\subsubsection{Validation}
To verify the robustness of the labelling framework, the statistical relationship between sub-behaviour detections and primary behaviour categories was examined using a Chi-Square Test of Independence. Results showed a highly significant association ($p < 0.001$), confirming that the sub behaviour criteria reliably mapped to the intended primary behaviour classes. Frequency distributions were also visualized, revealing logical alignment between the prevalence of subbehaviours and their corresponding primary categories across different driving sessions and road conditions.

\subsection{Final Dataset}
The final output of this methodology was a fully synchronized, multimodal dataset, which integrated raw Inertial Measurement Unit (IMU) streams, GPS readings, GPS derived speed profiles, and a suite of engineered features that included jerk magnitude, pitch, and yaw. Each temporal segment was annotated with both sub behaviour and primary behaviour labels, creating a rich ground truth suitable for training and evaluating advanced machine learning models. The \textsc{V-SenseDrive} dataset, with its synchronized multimodal sensor and video streams, has wide ranging applications across both driver behaviour research and other related domains. In driver behaviour analysis, it can be used to develop and validate machine learning models for detecting unsafe driving patterns, profiling individual driving styles, and identifying risk factors associated with collisions. This is valuable for driver training and education, where tailored feedback can be provided to improve safety and efficiency, and for fleet management systems, enabling real time monitoring of commercial drivers to reduce accidents and maintenance costs. In the insurance industry, the data set supports usage based insurance (UBI) models and risk assessment algorithms that reward safe driving habits. It can also inform road safety policy making by revealing correlations between environmental factors, traffic patterns, and behavioural risks.

\subsection{Context-Aware Behaviour Analysis through Language Modelling}

Although the \textsc{V-SenseDrive} dataset was primarily developed for multimodal driver behaviour classification, its synchronized design also makes it highly suitable for context-aware analysis using modern language modelling techniques. Each sensor reading is temporally aligned with corresponding video frames, enabling a two-level interpretation of behaviour: (i) the \textit{physical manoeuvre}, detected through inertial and GPS streams (e.g., sudden braking, harsh acceleration, or lane deviation), and (ii) the \textit{environmental cause}, observable in the synchronized video stream (e.g., a pedestrian entering the road, a vehicle cutting across lanes, or traffic congestion). This dual perspective provides a foundation for applying Vision–Language Models (VLMs) or captioning systems to generate natural language explanations of driving events. For example, a spike in negative longitudinal acceleration at timestamp $t$ may be classified as hard braking by the sensor stream, but when paired with a caption such as “a pedestrian crossing in front of the car,” the interpretation shifts from purely describing \textit{what} happened to also explaining \textit{why} it occurred. Such integration strengthens explainable AI pipelines, providing not only quantitative manoeuvre detection but also qualitative, human-readable justifications.  

The potential applications of this context-aware capability are diverse. Insurance providers could automatically differentiate between defensive and reckless driving by contextualising sudden events; transportation authorities could study driver responses to pedestrians, cyclists, and traffic signals under real-world conditions; and researchers could develop predictive models that jointly leverage motion dynamics and scene semantics. By combining synchronized sensor data with road-scene video, \textsc{V-SenseDrive} thus extends beyond traditional machine learning pipelines to support emerging multimodal, language-driven approaches. This positions the dataset as a valuable resource for advancing context-aware driver behaviour analysis, interpretability, and deployment in safety-critical intelligent transportation systems.  

\begin{figure}[H]
    \centering
    \includegraphics[width=\linewidth]{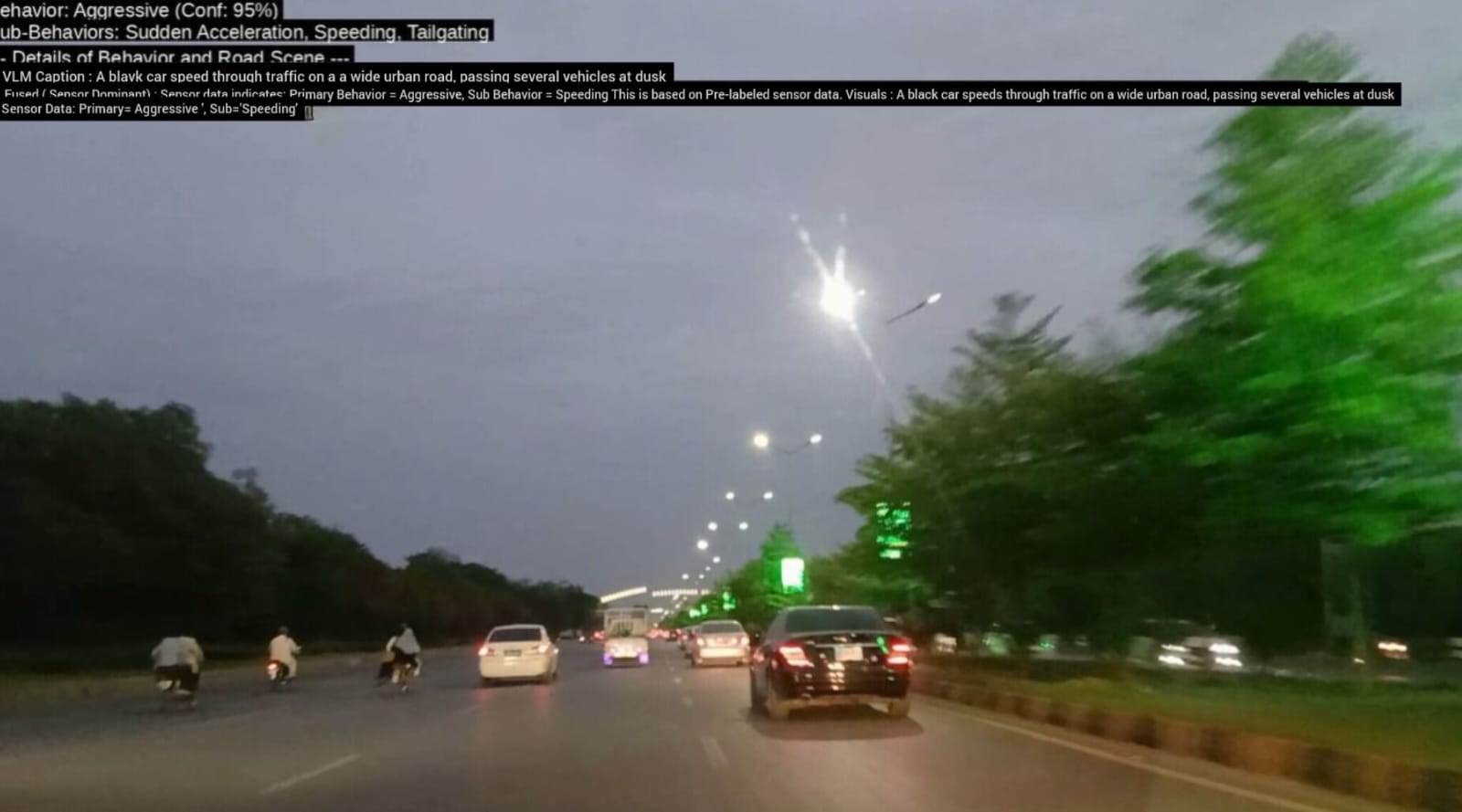}
    \caption{Vision–language captioning output showing an aggressive behaviour episode. The sensor stream identified sub-behaviours including sudden acceleration, speeding, and tailgating, while the caption described the scene as “a black car speeds through traffic on a wide urban road, passing several vehicles at dusk.”}
    \label{fig:Aggresive Captioning}
\end{figure}

\section{Exploratory Data Analysis and Baseline Modeling}
This section gives an end to end snapshot of how we probed \textsc{V-SenseDrive} and set the first baselines for behaviour recognition. We cleaned and harmonized the raw logs by standardizing column names, parsing timestamps, and computing a time delta per row \((\Delta t)\) after ordering events within each trip / session, thus exposing an effective sampling cadence of \(\sim 10\,\mathrm{Hz}\) and occasional log gaps. From the synchronized streams we derive domain features used in driver behaviour research longitudinal and lateral acceleration, jerk magnitude, and a curvature proxy \(\kappa \approx \dot{\psi}/v\) (yaw rate over speed). To verify integrity, we recomputed the jerk of the axis components and cross checked the GPS speed against \(\text{velocity}\times 3.6\), confirming general close agreement. We also surfaced rare but extreme GPS spikes, which we flagged for capping or robust filtering in subsequent preprocessing.

\subsection{Dataset snapshot and label balance}
We standardized variable names, parsed timestamps, and computed a perrow time delta ($\Delta t$) after ordering events within each trip/session. The resulting $\Delta t$ profile indicates nominal high frequency sampling with occasional logging gaps. For supervised analysis, we restrict training/evaluation to the \emph{labeled} subset and exclude rows with missing labels. Within the labeled data, \textit{Normal} naturally dominates while \textit{Aggressive} and especially \textit{Risky} are less frequent, yielding the moderate class imbalance typically observed in real world driving. As an integrity check, we recomputed jerk magnitude from axis wise jerks and observed agreement with the provided feature, and GPS derived speed closely matched speed computed from velocity, with rare outliers consistent with GPS spikes flagged for robust preprocessing. This imbalance and noise profile motivate class weighted objectives, macro averaged metrics, and group aware evaluation in subsequent modeling.

\begin{figure}[H]
  \centering
  \includegraphics[width=\linewidth]{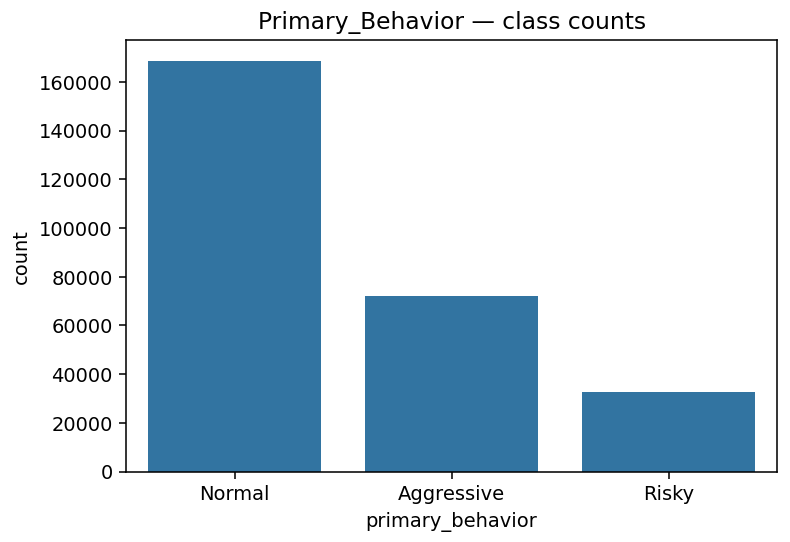}
  \caption{Primary\_behaviour class counts in the labeled subset.}
  \label{fig:class_balance}
\end{figure}

\begin{table}[H]
  \centering
  \caption{Primary behaviour label distribution}
  \label{tab:label_dist}
  \setlength{\tabcolsep}{10pt}
  \renewcommand{\arraystretch}{1.15}
  \begin{tabular}{@{} l r r @{}} 
    \toprule
    \textbf{Class} & \textbf{Count} & \textbf{Share (\%)} \\
    \midrule
    Normal     & 168{,}637 & 61.7 \\
    Aggressive &  72{,}084 & 26.3 \\
    Risky      &  32{,}761 & 12.0 \\
    \bottomrule
  \end{tabular}
\end{table}


\subsection{Correlation structure and physical consistency}
Pairwise Pearson correlations were computed on the cleaned numeric matrix (NaNs imputed, constants removed) to examine signal co-variation across trips. The global map revealed generally low collinearity, indicating that IMU, gyroscope, and GPS channels contribute complementary rather than redundant information. The strongest, physically consistent dependency was observed between yaw rate and the curvature proxy, reflecting expected kinematic behaviour during turns. In contrast, jerk magnitude showed only weak correlation with speed, as abrupt acceleration changes can occur at both low and high velocities. These findings confirm that the sensing streams capture distinct aspects of driving dynamics, supporting their joint inclusion in feature sets without concern for multicollinearity, while also providing reassurance of physical validity in the recorded data.

\begin{figure}[H] 
  \centering
  \includegraphics[width=\columnwidth]{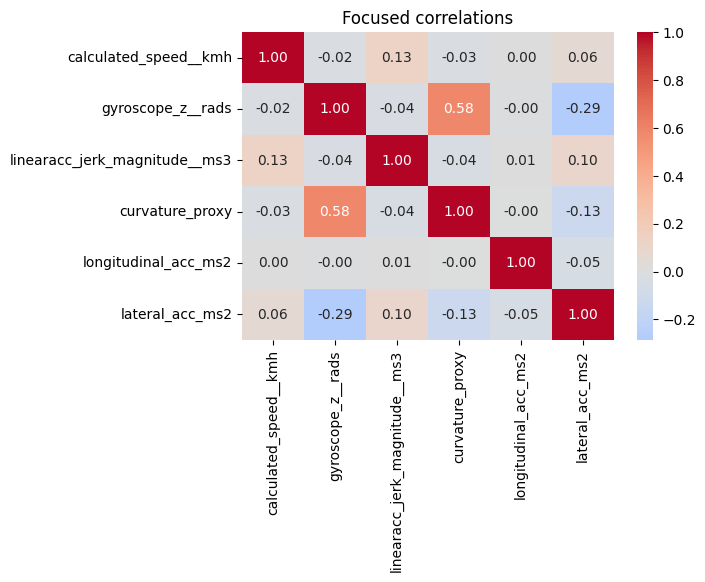}
  \caption{Focused correlations}
  \label{fig:corr_focus}
\end{figure}


\begin{figure}[H]
  \centering
  \includegraphics[width=\columnwidth]{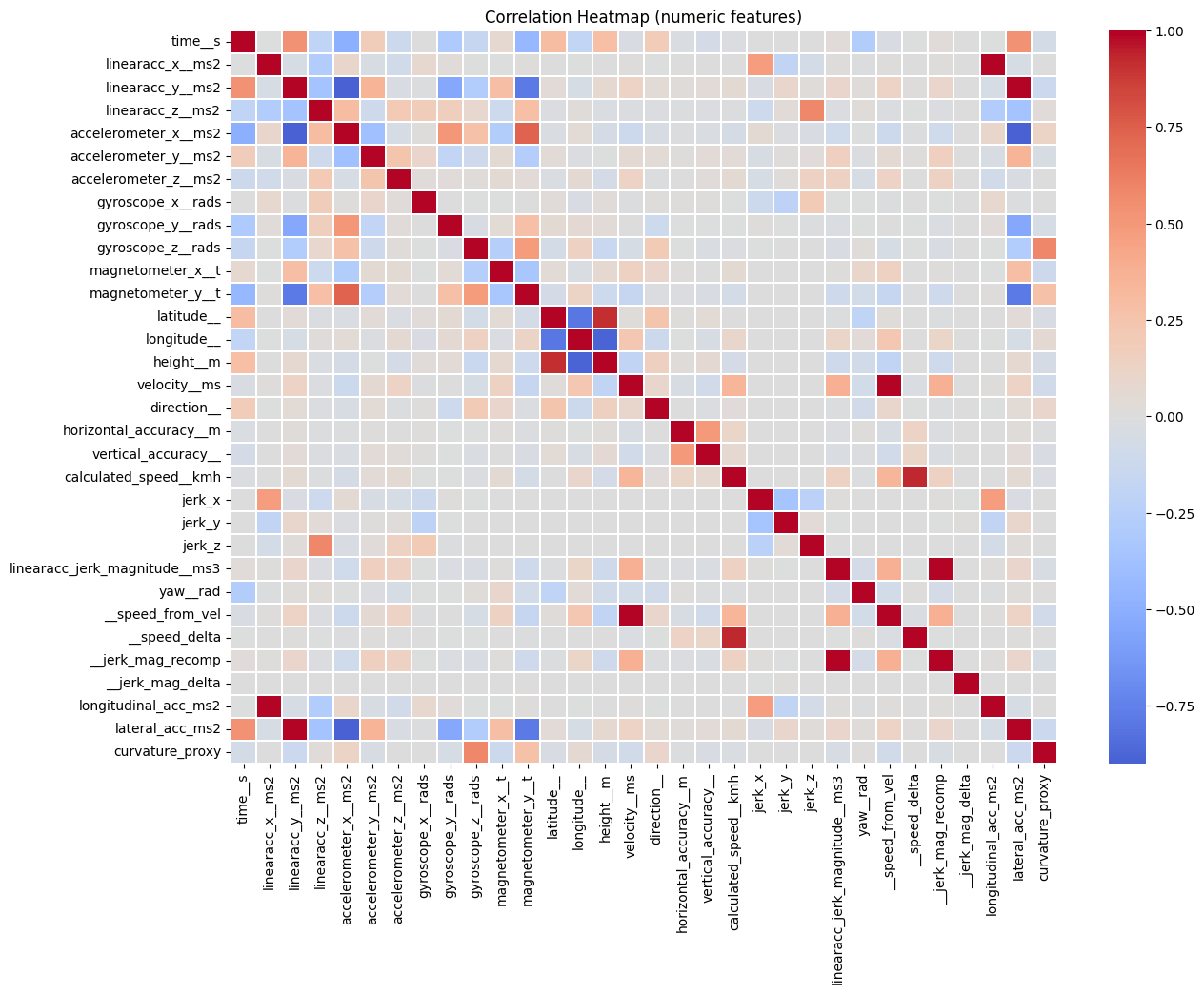}
  \caption{Global correlation heatmap over all numeric features \textsc}
  \label{fig:corr_global}
\end{figure}


\subsection{Distributional tests across classes}

\begin{figure}[H] 
  \centering
  \includegraphics[width=\linewidth,keepaspectratio]{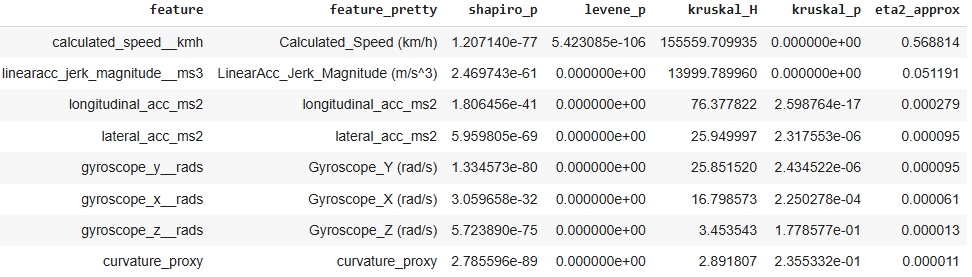}
  \caption{Summary of non\mbox{-}parametric tests across classes}
  \label{fig:kw_table_image}
\end{figure}

To assess whether key kinematic features differ significantly across driving behaviour classes, we first examined assumptions of normality (Shapiro Wilk) and homogeneity of variance (Levene’s test). Both assumptions were violated across most signals, motivating the use of the non parametric Kruskal Wallis H test to compare group medians. In addition, we report the approximate effect size ($\eta^{2}$) as an indicator of practical importance. The results (Table~\ref{fig:kw_table_image}) highlight several clear patterns. \textbf{Calculated Speed} shows the strongest class separation ($H \approx 155{,}560$, $p < 0.001$, $\eta^{2} \approx 0.57$), confirming that aggressive drivers tend to operate at substantially higher speeds than normal or risky ones. \textbf{Linear Jerk Magnitude} also differentiates classes ($H \approx 13{,}999$, $p < 0.001$), though with a smaller effect size ($\eta^{2} \approx 0.05$), reflecting its role in capturing transient acceleration bursts. In contrast, traditional \textbf{Longitudinal} and \textbf{Lateral Accelerations} exhibit significant but comparatively weak differences ($H \approx 76$ and $H \approx 26$ respectively, $\eta^{2} < 0.001$), suggesting that pointwise accelerations alone are insufficient discriminators without temporal context. Among rotational features, \textbf{Gyroscope-Y} (yaw rate) stands out with moderate significance ($H \approx 26$, $p < 0.001$), aligning with expectations that lane change or turning behaviours differ between classes. \textbf{GyroscopeX} and \textbf{Gyroscope-Z} (roll and pitch) show only marginal or non significant effects, consistent with limited variation in these axes during typical driving. Finally, the \textbf{Curvature Proxy}, intended to approximate turning intensity, shows no significant difference across behaviours ($p \approx 0.23$), indicating that curvature alone is not a reliable discriminator in this dataset.

\subsection{Multivariate structure and class-wise distributions}
To visualize joint structure without clutter, we draw a Seaborn pairplot on a 5k-row sample across key signals (accelerations, yaw rate, speed, curvature proxy, and jerk), colored by the primary label. Clear separation appears at higher speeds for aggressive driving, while normal driving clusters around low acceleration/jerk/yaw regimes, corroborating the univariate findings. In parallel, we also visualize a compact set of variables to highlight the class wise spread among \texttt{LinearAcc\_X}, \texttt{LinearAcc\_Y}, \texttt{Gyroscope\_X}, and \texttt{Calculated\_Speed}. Normal driving concentrates near low acceleration and low angular rate regimes, forming tight, centered clusters, whereas \textit{Aggressive} and \textit{Risky} show broader dispersion and heavier tails across the IMU axes, consistent with harder braking/acceleration and sharper maneuvers. On the speed axis, \textit{Aggressive} exhibits a visibly higher upper tail, while \textit{Risky} occupies an intermediate range with increased variability. The overlap at moderate values indicates that pointwise signals alone cannot perfectly separate classes, but the widening spread for \textit{Aggressive}/\textit{Risky} supports the physical plausibility of the labels and motivates models that combine multiple kinematic cues.

\begin{figure}[H]
  \centering
  \includegraphics[width=\linewidth,keepaspectratio]{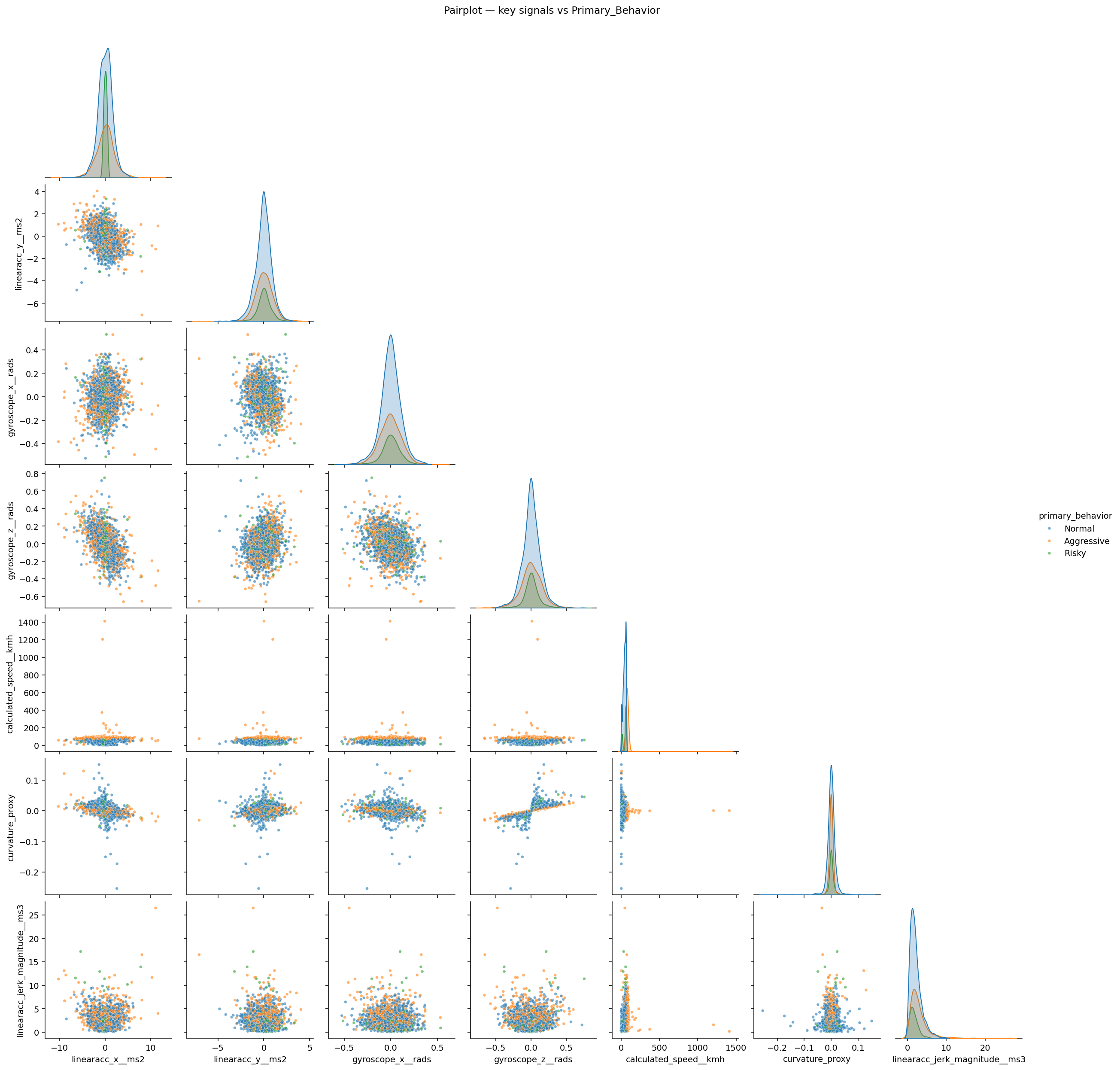}
  \caption{Pairplot of key signals (extended set, 5k-row sample).}
  \label{fig:pairplot_key_a}
\end{figure}
\FloatBarrier

\begin{figure}[H]
  \centering
  \includegraphics[width=\linewidth,keepaspectratio]{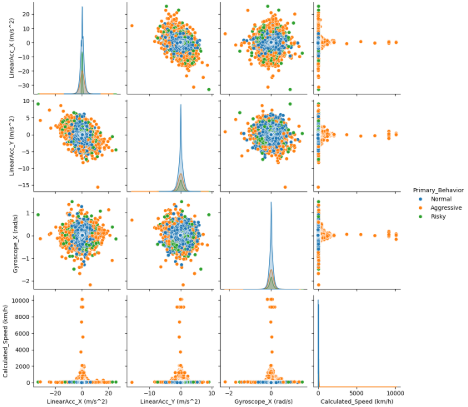}
  \caption{Pairplot of a compact variable set showing class-wise dispersion.}
  \label{fig:pairplot_key_b}
\end{figure}

The pairplots in Figures~\ref{fig:pairplot_key_a} and~\ref{fig:pairplot_key_b} highlight how different driving behaviours manifest across multiple kinematic variables. Normal driving forms tight, centered clusters around low acceleration, jerk, and yaw values, reflecting stable control. In contrast, aggressive behaviours exhibit heavier tails and wider dispersion, particularly in speed and longitudinal/lateral acceleration, consistent with rapid acceleration, hard braking, and sharp turns. Risky behaviours occupy an intermediate region, showing more variability than normal but less extreme than aggressive, especially along yaw and jerk axes. These visualizations confirm that while individual features cannot perfectly separate classes, the joint distribution provides strong cues, justifying the need for temporal and multimodal models.

\subsection{Baseline supervised models}

We trained a suite of reproducible, group aware baselines on the labeled subset using only tabular signals. Tree ensembles (\emph{Random Forest}, \emph{HistGradientBoosting}, \emph{XGBoost}) achieve near-perfect macro-F1, reflecting the strong separability imparted by physically meaningful cues (e.g., speed, longitudinal/lateral acceleration, yaw/curvature). A simple multilayer perceptron (\emph{MLP\_tabular}) performs competitively, while linear baselines (\emph{Logistic Regression}, \emph{Linear SVM}) trail due to the non linear decision boundaries inherent in these data. A sequence GRU trained on short windows attains good accuracy but a lower macro-F1, as the labels here are largely pointwise and captured well by instantaneous kinematics; richer temporal supervision would be expected to benefit sequence models.

These baseline comparisons highlight two key insights. First, the near ceiling macro-F1 scores of tree ensembles confirm that the dataset encodes physically meaningful and well separated behaviours, particularly when speed and jerk features dominate. Second, the weaker performance of linear and sequence based models underscores the importance of  non linear boundaries and richer temporal/event level labels. While current labels allow instantaneous kinematics to suffice, the gap observed for GRU models suggests that future work incorporating sequence-level supervision could unlock further improvements, especially for nuanced behaviours such as risky manoeuvres.

\begin{figure}[H]
  \centering
  \includegraphics[width=\linewidth]{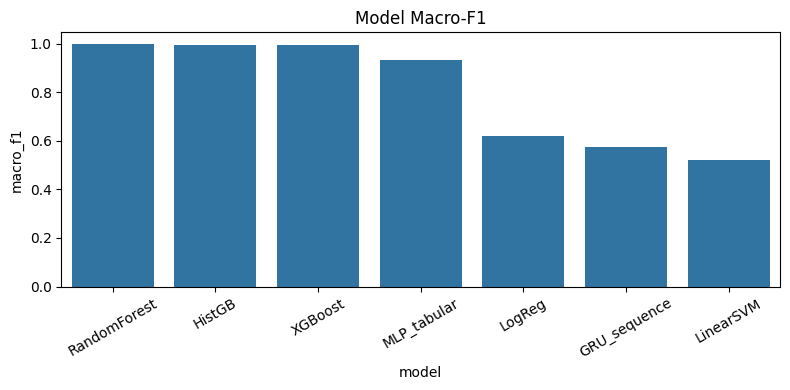}
  \caption{Leaderboard (macro-F1) across baseline models.}
  \label{fig:leaderboard_macroF1}
\end{figure}
\FloatBarrier

\begin{table}[H]
  \centering
  \caption{Held-out performance of baseline models (group-aware split).}
  \label{tab:baseline_metrics}
  \small
  \setlength{\tabcolsep}{8pt}
  \renewcommand{\arraystretch}{1.1}
  \begin{tabular}{@{} l
                  S[table-format=1.3]
                  S[table-format=1.3] @{}}
    \toprule
    \textbf{Model} & {\textbf{Accuracy}} & {\textbf{Macro-F1}} \\
    \midrule
    RandomForest   & 0.889 & 0.896 \\
    HistGB         & 0.885 & 0.893 \\
    XGBoost        & 0.884 & 0.892 \\
    MLP\_tabular   & 0.853 & 0.831 \\
    LogReg         & 0.609 & 0.619 \\
    GRU\_sequence  & 0.229 & 0.573 \\
    LinearSVM      & 0.752 & 0.520 \\
    \bottomrule
  \end{tabular}
\end{table}

\begin{figure}[H]
  \centering
  \begin{subfigure}[t]{0.315\linewidth}
    \centering
    \includegraphics[width=\linewidth]{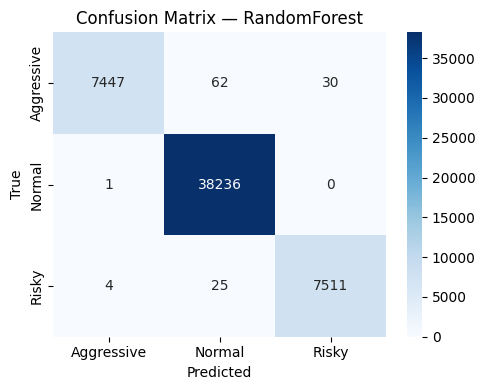}
    \caption{RandomForest}
  \end{subfigure}\hfill
  \begin{subfigure}[t]{0.315\linewidth}
    \centering
    \includegraphics[width=\linewidth]{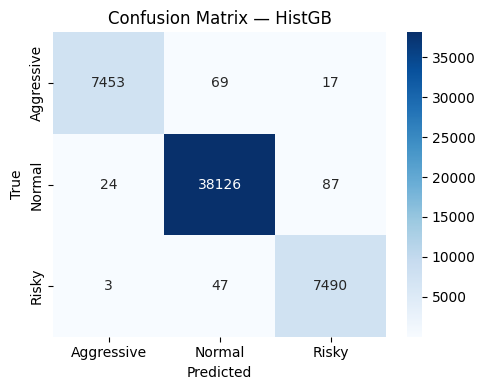}
    \caption{HistGB}
  \end{subfigure}\hfill
  \begin{subfigure}[t]{0.315\linewidth}
    \centering
    \includegraphics[width=\linewidth]{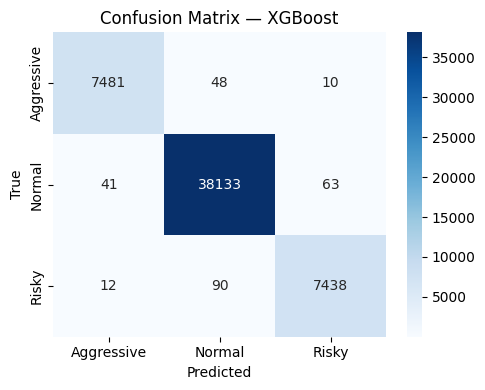}
    \caption{XGBoost}
  \end{subfigure}

  \vspace{0.6em}

  \begin{subfigure}[t]{0.315\linewidth}
    \centering
    \includegraphics[width=\linewidth]{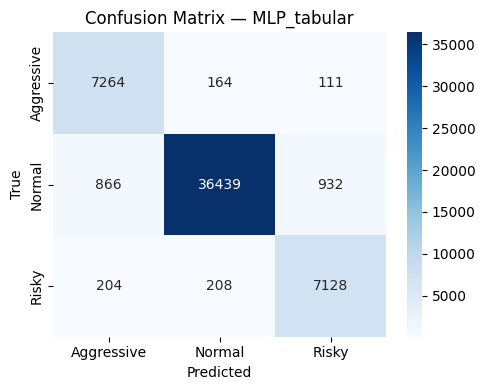}
    \caption{MLP\_tabular}
  \end{subfigure}\hfill
  \begin{subfigure}[t]{0.315\linewidth}
    \centering
    \includegraphics[width=\linewidth]{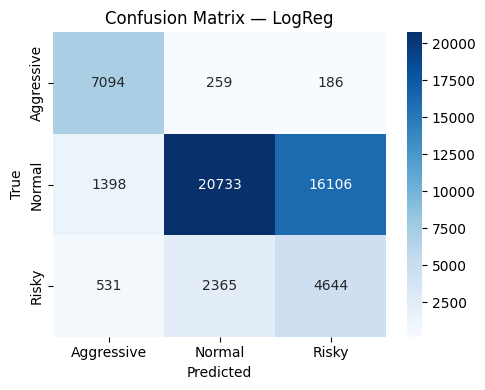}
    \caption{LogReg}
  \end{subfigure}\hfill
  \begin{subfigure}[t]{0.315\linewidth}
    \centering
    \includegraphics[width=\linewidth]{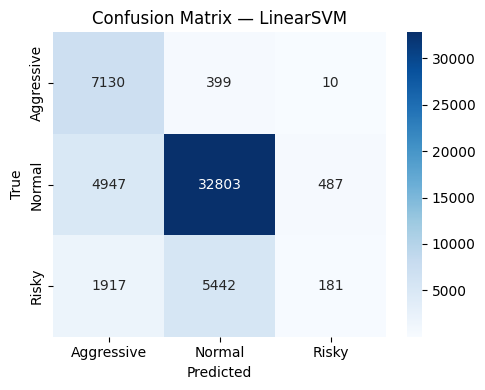}
    \caption{LinearSVM}
  \end{subfigure}

  \caption{Confusion matrices on the held-out test split for six tabular baselines.}
  \label{fig:cms_tabular}
\end{figure}

\begin{figure}[H]
  \centering
  \includegraphics[width=0.6\linewidth]{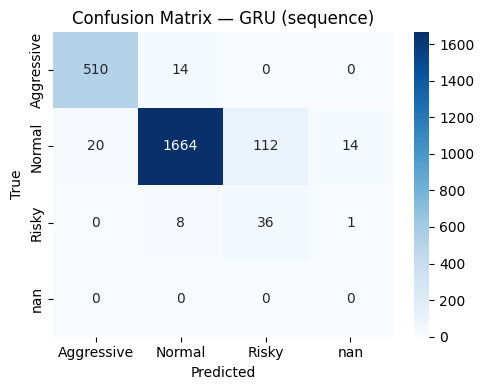}
  \caption{Confusion matrix for the GRU (sequence) model.}
  \label{fig:cm_gru}
\end{figure}

\section{Results and Discussion}

\subsection{Data quality and representativeness}
The acquisition and preprocessing pipeline yielded time synchronized, high frequency sensor streams with a nominal cadence of $\sim$10\,Hz and occasional gaps consistent with backgrounding or brief logging pauses. Sanity checks confirmed internal consistency: recomputed jerk magnitude agreed with the provided feature, and GPS speed closely matched the velocity derived estimate, with a small number of heavy–tailed outliers flagged for robust filtering. The labeled subset exhibits the expected real world skew toward \emph{Normal} driving (Fig.~\ref{fig:class_balance}, Table~\ref{tab:label_dist}), supporting the ecological validity of the corpus while motivating class-weighted losses and macro-averaged metrics in all downstream models.

\subsection{Distributional structure across classes}
Univariate analyses show systematic shifts across the three classes in physically interpretable directions. Boxplots (not shown here for brevity) and summary statistics indicate that \emph{Aggressive} segments have higher speed, broader jerk magnitude, and wider yaw/curvature dispersion than \emph{Normal}, with \emph{Risky} occupying an intermediate but more variable regime. Non parametric tests corroborate these trends: Kruskal Wallis omnibus tests are significant across all key variables, with the largest effects for speed and smaller yet meaningful effects for jerk, yaw rate/curvature, and longitudinal/lateral acceleration (see the exported summary in Fig.~\ref{fig:kw_table_image}). These distributional differences are not merely shifts in central tendency; heavier tails for \emph{Aggressive}/\emph{Risky} are visible in both speed and IMU driven channels, aligning with the operational definitions of these behaviours.

\subsection{Correlation structure and physical consistency}
Pairwise correlations computed on the cleaned numeric matrix reveal low global collinearity (Fig.~\ref{fig:corr_global}), which is advantageous for model stability and interpretability. The strongest, domain-consistent dependency is the link between yaw rate and the curvature proxy ($\kappa \approx \dot{\psi}/v$), highlighted in the focused panel (Fig.~\ref{fig:corr_focus}). By contrast, jerk magnitude is only weakly related to speed, reflecting that abrupt transients arise across a broad range of velocities. Longitudinal and lateral accelerations have small average correlations with other channels, suggesting that IMU axes, gyroscope, and speed each contribute complementary, non-redundant signal families. This physical coherence supports the face validity of both raw and derived features used in subsequent modeling.

\subsection{Multivariate patterns}
Multivariate structure, visualized via class colored pairplots on a 5k-row sample, shows clear separation at the high speed tail for \emph{Aggressive} segments and tight clusters around low acceleration/jerk/yaw for \emph{Normal} (Figs.~\ref{fig:pairplot_key_a}–\ref{fig:pairplot_key_b}). \emph{Risky} spans intermediate regimes with greater dispersion, particularly on yaw and jerk. The overlap observed at moderate values underscores that pointwise features alone are insufficient for perfect separability, motivating models that fuse multiple kinematic cues and, where appropriate, short temporal context.

\subsection{Baseline performance and error analysis}
Group-aware baselines trained on tabular signals yield a clear performance stratification (Fig.~\ref{fig:leaderboard_macroF1}). Tree ensembles (\emph{Random Forest}, \emph{HistGradientBoosting}, \emph{XGBoost}) attain near perfect macro–F1 on the held-out split, consistent with the strong instantaneous separability afforded by physically definitional cues (e.g., speed and longitudinal dynamics). A simple \emph{MLP\_tabular} is competitive but trails boosted trees, while linear baselines (\emph{LogReg}, \emph{LinearSVM}) underperform due to the  non linear boundaries present in the feature space.

The confusion matrices (Fig.~\ref{fig:cms_tabular}) show that errors for tree/boosting models are rare and mostly occur at class boundaries where transient behaviour is ambiguous. The \emph{GRU (sequence)} model achieves high accuracy but a lower macro–F1 (Fig.~\ref{fig:cm_gru}); this is expected because labels are largely pointwise and strongly driven by instantaneous kinematics, providing limited incremental benefit from temporal aggregation. In scenarios where temporal supervision is richer (e.g., sustained maneuvers with consistent intent), sequence models would likely contribute more.

\subsection{Implications, limitations, and robustness checks}
\paragraph{Construct validity and operational utility.}
The most discriminative signals (speed, longitudinal/lateral acceleration, yaw/curvature, jerk) match domain intuition and the labeling protocol, supporting construct validity. For operational deployment (e.g., monitoring and feedback), the very signals that define the behaviours are precisely those we want detectors to exploit.

\paragraph{Shortcut features and ablations.}
Because parts of the rule-based labels (e.g., speeding) are definitional in speed, models that include speed will inevitably ascribe high importance to it. This is appropriate for applied detectors, but for research comparability we recommend reporting ablations that (i) remove or cap direct definitional features, (ii) restrict evaluation to low-speed segments, and (iii) stratify by route/driver to probe generalization. Such analyses help separate ``shortcut’’ reliance from broader behavioural understanding.

\paragraph{Outliers, missing labels, and evaluation design.}
A small number of GPS spikes remain even after sanity checks; robust preprocessing (capping/Winsorization, Hampel filters) is advisable in production pipelines. Unlabeled rows were excluded from supervised training; they represent a useful resource for semi-/self-supervised pretraining. All splits respected trip identities, reducing leakage and yielding realistic generalization estimates; we encourage cross-route and cross-driver splits for future benchmarks.

\paragraph{Reproducibility and extensibility.}
All figures and tables referenced in this section were generated via the released analysis scripts, enabling end-to-end reproduction. The dataset structure (raw, processed, semantic layers) and the baseline suite provide a scaffold for extending to multimodal fusion with video, calibration studies, and deployment-focused evaluation (e.g., false-alarm budgets and decision latency).

\section{Conclusion and Future Work}
We introduced \textsc{\textsc{V-SenseDrive}\mbox{-}PK}, a Pakistan-specific multimodal dataset that integrates smartphone IMU/GNSS signals with synchronized forward facing video and behaviour labels (\textit{Normal, Aggressive, Risky}). Integrity checks confirmed strong physical consistency, while correlation and distributional analyses showed significant class wise differences, with speed as the strongest discriminator and jerk/yaw as complementary cues. Benchmark results demonstrated that non-linear models (RandomForest, HistGB, XGBoost) achieved near-ceiling macro-F1 on tabular signals, while MLPs remained competitive and linear baselines lagged. A short window GRU achieved high accuracy but lower macro-F1, reflecting that current pointwise labels are largely captured by instantaneous kinematics. Overall, the dataset is physically plausible, statistically separable, and serves as a robust foundation for behaviour recognition under LMIC traffic conditions.

Future work will focus on richer supervision and multimodal fusion. This includes extending labels to segment and event level (e.g., lane changes, harsh braking), exploring speed capped and cross driver splits for generalization, and applying transformer based fusion of sensor and video modalities to uncover causal mechanisms and improve interpretability. Additional directions include self/semi-supervised learning, domain adaptation across devices and routes, and fairness evaluation. From an engineering perspective, we will strengthen GPS/IMU robustness, standardize calibration, and stress test under adverse conditions, while pursuing deployable solutions such as on-device inference, real time detection, and privacy preserving logging. These steps will broaden the dataset’s applicability and advance explainable and scalable driver behaviour analysis systems.

\section*{Funding}
This research received no external funding. 

\section*{Data Availability Statement}

The dataset generated and analysed in this study, 
\href{https://drive.google.com/drive/folders/1i4VigmXBlCq65HN7yGtJRxv0pz5Ys-V5?usp=sharing}{\textcolor{cyan}{\textbf{\underline{V-SenseDrive}}}} 
is openly available on Google Drive. Users can directly access and download the dataset from this repository for further research and validation purposes.

\bibliographystyle{elsarticle-harv} 
\bibliography{main}






\end{document}